\documentclass[11pt,a4paper]{article}
\usepackage[hyperref]{emnlp2020}
\usepackage{times}
\usepackage{latexsym}

\usepackage{microtype}

\aclfinalcopy % Uncomment this line for the final submission
\def\aclpaperid{2999} %  Enter the acl Paper ID here

\usepackage{url}
\usepackage{soul}
\usepackage{graphicx}
\usepackage[toc,page]{appendix}
\usepackage[export]{adjustbox}
\usepackage{amsmath, amsfonts}
\usepackage{graphicx}
\usepackage{pifont}
\usepackage{booktabs}
\usepackage{multirow}
\newcommand{\cmark}{\ding{51}}
\newcommand{\xmark}{\ding{55}}
\usepackage{rotating}
\usepackage{subfig}
\title{Precise Task Formalization Matters in Winograd Schema Evaluations}
\author{Haokun Liu$^1$\thanks{~~Equal contribution.} ~~William Huang$^2$\footnotemark[1] ~~Dhara A. Mungra$^1$ ~~Samuel R. Bowman$^{1,2,3}$\\
$^1$Center for Data Science, New York University\\
$^2$Courant Institute of Mathematical Sciences, New York University\\
$^3$Department of Linguistics, New York University\\
\texttt{\{haokunliu, wh629, dam797, bowman\}@nyu.edu}
}

\begin{document}

\maketitle
\begin{abstract}
Performance on the Winograd Schema Challenge (WSC), a respected English commonsense reasoning benchmark, recently rocketed from chance accuracy to 89\% on the SuperGLUE leaderboard, with relatively little corroborating evidence of a correspondingly large improvement in reasoning ability. We hypothesize that much of this improvement comes from recent changes in task formalization---the combination of input specification, loss function, and reuse of pretrained parameters---by users of the dataset, rather than improvements in the pretrained model's reasoning ability. We perform an ablation on two Winograd Schema datasets that interpolates between the formalizations used before and after this surge, and find (i) framing the task as multiple choice improves performance by 2-6 points and (ii) several additional techniques, including the reuse of a pretrained language modeling head, can mitigate the model's extreme sensitivity to hyperparameters. We urge future benchmark creators to impose additional structure to minimize the impact of formalization decisions on reported results.
\end{abstract}

\begin{table*}[t]
    \centering
    \resizebox{.99\textwidth}{!}{
    \begin{tabular}{p{.17\textwidth} p{.15\textwidth} p{.37\textwidth}  p{.07\textwidth} l  p{.03\textwidth} p{.07\textwidth}}
        \toprule
        Ablation & Formalization & Input Example & Emb & Loss & MC & Label\\
        \midrule
        \multirow{2}{*}{\shortstack[l]{Strong Baseline\\
        (RoBERTa\\ method)}} & \multirow{2}{*}{MC-MLM} & a) \texttt{[CLS]} \underline{Jim} yelled at Kevin because   \texttt{[MASK]} was so upset. \texttt{[SEP]} & \multirow{2}{*}{\texttt{[MASK]}} &$\begin{aligned}-\sum_i^n \delta_{y,i} \log \frac{\mathbb{P}(\text{NP}_i\vert \text{s})}{\sum_j^n\mathbb{P}(\text{NP}_j\vert \text{s})}\end{aligned}$ & \multirow{2}{*}{\cmark} & \multirow{2}{*}{Index} \\
        & & b) \texttt{[CLS]} Jim yelled at \underline{Kevin} because   \texttt{[MASK]} was so upset. \texttt{[SEP]} & & & & \\
        \midrule
        \multirow{2}{*}{\shortstack[l]{No MLM\\
        (WinoGrande\\ method)}} & \\
        & MC-Sent & \multirow{8}{*}{\shortstack[l]{\vspace{12mm} \\ a) \texttt{[CLS]} Jim yelled at Kevin because \\ Jim \texttt{[SEP]} was so upset. \texttt{[SEP]}\\\vspace{10mm} \\ b) \texttt{[CLS]} Jim yelled at Kevin because \\ Kevin \texttt{[SEP]} was so upset. \texttt{[SEP]}}} & \multirow{8}{*}{\shortstack[l]{\vspace{15mm} \\\texttt{[CLS]}}} & $\begin{aligned}-\sum_i^n \delta_{y,i} \log \mathbb{P}(\text{NP}_i\vert \text{s}_1,\ldots, \text{s}_n)\end{aligned}$ & \cmark & Index \\
        &\\
        % with multi-line equations
        \shortstack[l]{No Softmax \\ Scaling} & MC-Sent-NoSoftmax & & & $\begin{aligned}-\sum_i^n \big[&y_i \log \mathbb{P}(\text{True}\vert \text{s}_i) \\& + (1-y_i) \log \mathbb{P}(\text{False}\vert \text{s}_i)\big]\end{aligned}$ & \cmark & Binary \\
        &\\
        \shortstack[l]{No Paired \\ Training} & MC-Sent-NoPairLoss & & & $\begin{aligned}-\big[&y \log \mathbb{P}(\text{True}\vert \text{s})\\ &+ (1-y) \log \mathbb{P}(\text{False}\vert \text{s})\big]\end{aligned}$ & \cmark & Binary \\
        &\\
        \shortstack[l]{No MC \\ Evaluation} & P-Sent & & & $\begin{aligned}-\big[&y \log \mathbb{P}(\text{True}\vert \text{s})\\ &+ (1-y) \log \mathbb{P}(\text{False}\vert \text{s})\big]\end{aligned}$ & \xmark & Binary \\
        &\\
        \midrule
        \multirow{1}{*}{\shortstack[l]{Predict from Span\\ (SuperGLUE\\ method)}} & P-Span  & \texttt{[CLS]} \underline{Jim} yelled at Kevin because \underline{he} was so upset. \texttt{[SEP]} & \texttt{[CLS]}, \textsc{Pron}, \text{NP} & $\begin{aligned}-\big[&y \log \mathbb{P}(\text{True}\vert \text{s})\\ &+ (1-y) \log \mathbb{P}(\text{False}\vert \text{s})\big]\end{aligned}$ & \xmark & Binary \\
        \bottomrule
    \end{tabular}}
    \caption{Overview of the formalizations. When \textit{MC} (multiple choice) is \cmark, the model predicts positive if the query NP $\mathbb{P}(\cdot)$ is highest among all candidates; when \textit{MC} is \xmark, the model predicts positive if $\mathbb{P}(\cdot)>0.5$. \textit{Emb} indicates which RoBERTa output layer embeddings are used. In the loss function, $y$ is the index of the correct input when \textit{Label} is \textit{Index} and 0 or 1 when \textit{Label} is \textit{Binary}. \textit{s} is a sequence of input tokens, we use subscript to indicate multiple input sequences. $\delta_{y,i}$ is 1 when $y=i$ i.e. the i-th input is correct and 0 otherwise. In the MC-MLM input example, the underline marks NPs to predict. For P-Span, the underline marks the NP and \textsc{Pron} spans.}
    \label{tab:formalizations}
\end{table*}

\section{Introduction}
\label{sec:intro}

Over the last couple of years, large pretrained models have achieved human performance on a large share of established natural language understanding benchmark datasets \citep{devlin2018BERT}. Recent results report a surge in performance to near-human levels on the Winograd Schema Challenge \citep[WSC;][]{levesque2011WSC} in particular\citep{liu2019RoBERTa}. However, variations in task formulation across papers and evaluations makes it hard to understand the true degree of recent progress.

The WSC is an English commonsense reasoning evaluation that requires a model to resolve carefully-constructed ambiguous pronouns. For example, in the sentence ``\textit{\underline{Jim} yelled at Kevin because \textbf{he} was so upset.}'' the reader will likely have to consider the motivation of the query \underline{noun phrase} (NP) to recognize whether the pronoun \textit{\textbf{he}} refers to \textit{\underline{Jim}}.

The accuracy of WSC has seen an abrupt increase from 64\% to 89\% on the SuperGLUE \citep{wang2019superglue} leaderboard upon the release of RoBERTa \citep{liu2019RoBERTa}. While many works \citep{kocijan2019WSCTrick, liu2019RoBERTa, raffel2019t5} attribute such improvements to improved pretraining and the use of auxiliary training datasets, the impact of the task formalization---the combination of input specification, task specific layer design, and loss function---has not yet been seriously studied.

The SuperGLUE WSC baseline with BERT (64\%) resolves pronoun references for individual examples by concatenating the pronoun and query NP embeddings and making a binary prediction for the NP span. Meanwhile, RoBERTa (89\%) uses a pretrained masked language modeling (MLM) head as part of the output layer and treats the task as a multiple-choice (MC) decision between candidate NPs. We refer to these two task formalizations as pointwise span (\textbf{P-Span}) and multiple choice masked language modeling (\textbf{MC-MLM}).

In our work, we interpolate between P-Span and MC-MLM using both BERT and RoBERTa to understand tasks' sensitivity to formalization and the components contributing to MC-MLM's improvement. We find MC-MLM outperforms P-Span and reduces sensitivity to hyperparameters and random restarts. We also see large variances of scores spanning random guessing to state-of-the-art (SotA) performance. The biggest gain comes from including MC inference. Further, paired training with query and candidate NPs, using a softmax over candidates, and using a pretrained MLM head all lead to reductions in variance. We show that these formalization choices impact performance differences between the BERT and RoBERTa approaches on SuperGLUE WSC, with validation accuracy increasing between P-Span and MC-MLM by 21.1\% using RoBERTa and 10.5\% using BERT.

The effect of task formalization may incentivize gains from supplemental MC options and aggressive hyperparameter tuning. To avoid this, we suggest future benchmarks impose more structure, such as in this case either explicitly distributing gold candidate NPs or enforcing rules against their use. For system developers, this result highlights the value of fine-tuning pretrained language modeling heads to target tasks in low-resource settings \citep{raffel2019t5}, at least where the task format makes this an option.

\section{Related Work}
\label{sec:related}

\paragraph{WSC Datasets}
\citet{levesque2012WSC} launch the WSC with 108 handbuilt question-answer pairs, which has since grown to 273 examples, often called WSC273. Since then, several similar or derived datasets have emerged \citep{kocijan2020review}. The SuperGLUE version of the task recasts examples from WSC273 and the Pronoun Disambiguation Problems dataset \citep{morgenstern2016pdp} into 554 training and 104 validation binary classification problems. 146 test examples are derived from fiction books and handcrafted by the original WSC authors. \citet{sakaguchi2019winogrande} collect a larger dataset of fill-in-the-blank-format problems, called WinoGrande, via crowdsourcing with adversarial filtering. The dataset has five training sets, ranging from 160 to 41k examples, and shared validation and test sets with 1.3k and 1.8k examples, respectively. 

\paragraph{Approaches}
\citet{trinh2018lmWSC} generate inputs for their recurrent neural network language model by replacing the \textsc{Pron} with either the query or candidate NP and compare the probability of the two sentences, yielding 64\% accuracy on WSC273. \citet{radford2019gpt2} use this method with a transformer language model, boosting accuracy to 71\%. \citet{ruan2019bertWSC} fine-tune BERT in a similar way, reaching 71\% as well. \citet{kocijan2019WSCTrick} also use BERT, but include additional Winograd-like training data and use the model's pretrained MLM head to achieve 74\% accuracy. In another style of approach, \citet{klein2019attention} experiment with BERT pretrained \textit{attention} weights without fine-tuning, and achieve an accuracy of 60\%.

For SuperGLUE WSC, the official baseline uses BERT and a linear classifier on BERT's output embeddings for the \texttt{[CLS]} token, pronoun token, and query NP span representations but fail to exceed the majority-class baseline, only matching it at 64\%. \citet{liu2019RoBERTa} use the newer RoBERTa and adapt the \citet{kocijan2019WSCTrick} approach with cross entropy loss to raise this accuracy to 89\%. T5 \citep{raffel2019t5} marks the pronoun in the input and fine-tune a transformer encoder-decoder model to generate the target NP, achieving the current state of the art at 94\%.

Looking to WinoGrande, \citet{sakaguchi2019winogrande} adapt \citet{ruan2019bertWSC}'s method with RoBERTa as the baseline model, achieving 68\% accuracy on WinoGrande-Medium and 79\% accuracy on the full test set.

\section{Methods under Study}
\label{sec:method}
We evaluate six formalizations---three existing ones and three that we introduce---to interpolate between P-Span and MC-MLM. These all use an output layer on top of an MLM pretrained transformer model, but differ in the input specification, loss function, prediction method, contextual embeddings used by the output layer, and label type. Table \ref{tab:formalizations} presents an overview.

\paragraph{MC-MLM} This approach follows that of \citet{liu2019RoBERTa} in the introduction of RoBERTa. Here, the pronoun in the input is replaced by \texttt{[MASK]}. The model then uses its pretrained MLM head to evaluate the probability $\text{NP}_i$ should replace \texttt{[MASK]} and uses a softmax over the log probabilities. For multi-token NPs the model compares the \textit{geometric mean} of these probabilities.

\paragraph{MC-Sent} This approach follows the WinoGrande baselines. Here, we specify the inputs by replacing the pronoun with an NP candidate and marking it with an additional \texttt{[SEP]} token. The output head feeds each option's \texttt{[CLS]} embedding into a linear layer and applies a softmax over the outputs. MC-Sent trains a linear layer from scratch, while MC-MLM may take advantage of the embedding model's MLM pretraining.% Comparing the two allows us to evaluate the impact of using pretrained components.

\paragraph{MC-Sent-NoSoftmax} MC-Sent-NoSoftmax only differs from MC-Sent by replacing the final softmax with a sigmoid and computes the probabilities of whether each input sequence is correct. Without softmax, MC-Sent-NoSoftmax is unable to provide larger gradients for examples with smaller margins between candidates. We refer to this as softmax scaling.

\paragraph{MC-Sent-NoPairLoss} MC-Sent-NoPairLoss and MC-Sent-NoSoftmax differ by loss function, where MC-Sent-NoPairLoss only considers the query input. MC-Sent-NoPairLoss is unable to use gradients from multiple candidates to neutralize signals from shared words and focus on NP options. We refer to this as paired training.

\paragraph{P-Sent} In P-Sent, we further remove \textit{MC evaluation} by restricting the model to a single binary classification question. This forces P-Sent to resolve pronoun references without implicitly learning to detect and eliminate NPs.

\paragraph{P-Span} Instead of replacing \textsc{Pron} with NPs to determine the validity of the input sentence, P-Span follows the SuperGLUE baseline to determine whether the NP reference is correct. It first averages over the representations from the \textsc{Pron} and NP spans to create span representations. The span representations are then concatenated with the \texttt{[CLS]} token embedding and used by a logistic regression classifier.

\begin{figure*}[t]
    \centering
    \subfloat{
            \includegraphics[width=1\textwidth]{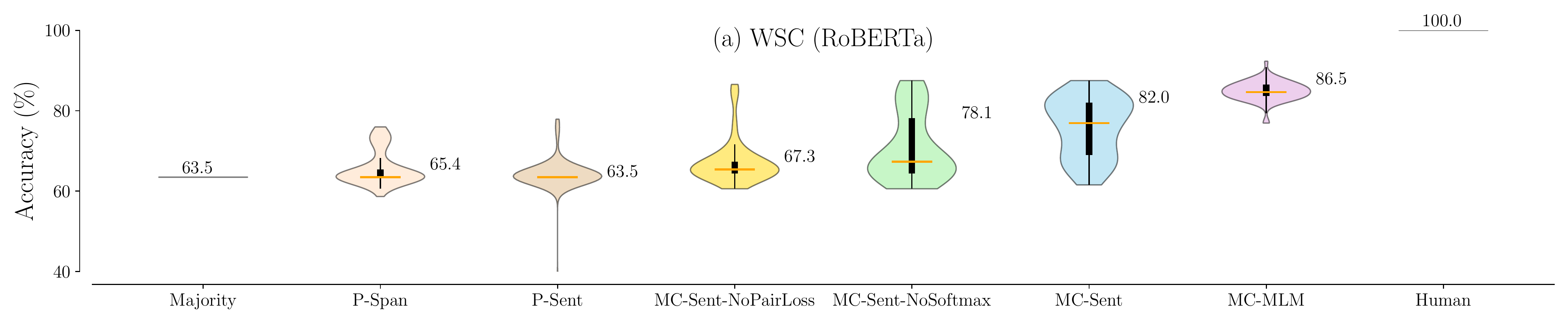}
            \label{subfig:wsc}
        }
        \vspace{-5mm}
        
    \subfloat{
            \includegraphics[width=1\textwidth]{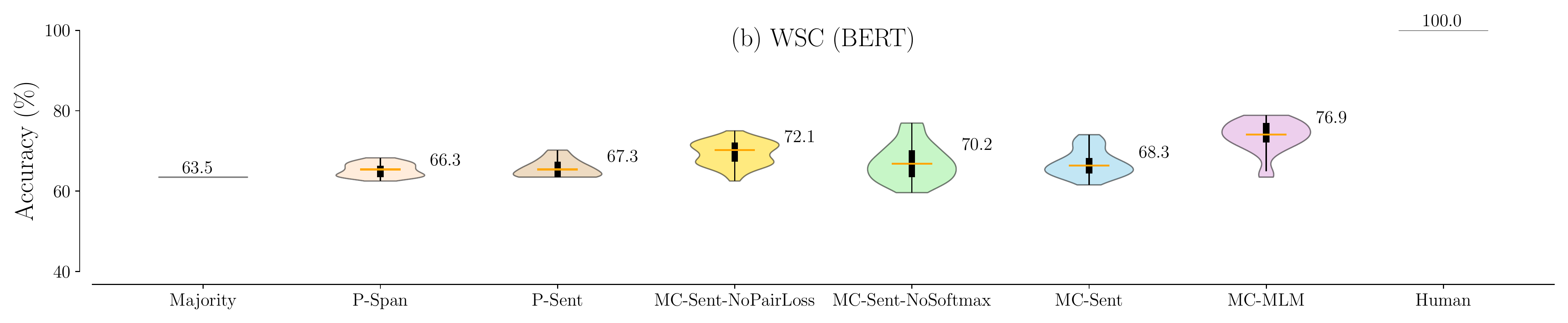}
            \label{subfig:bert_results}
        }
        \vspace{-5mm}
    
    \subfloat{
            \includegraphics[width=1\textwidth]{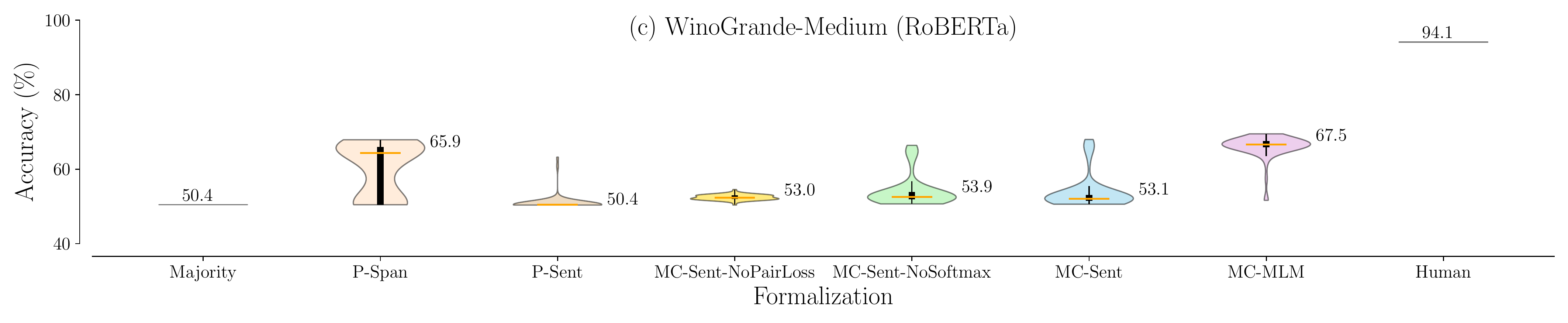}
            \label{subfig:winogrande}
        }
    \caption{Plots of validation accuracy from 60 runs on each corpus. The orange line marks the median number and label marks 75\textsuperscript{th} percentiles.}
    \label{fig:results}
\end{figure*}

\section{Experiments}
\label{sec:experiments}

\paragraph{Implementation}
Our code\footnote{\url{https://github.com/nyu-mll/wsc-formalizations/tree/code_release}} builds on Huggingface Transformers \citep{wolf2019huggingface} and fairseq \citep{ott2019fairseq}. All our experiments use either pretrained RoBERTa-large or BERT-large-cased models. 
% We follow  \citeauthor{liu2019RoBERTa}, we use AdamW \cite{loshchilov2018adamw} optimizer with (0.9, 0.999) $\beta$'s and 0.001 weight decay. We use a linear learning rate scheduler with 6\% warm up steps. 
We evaluate on the validation set every epoch with early stopping. We conduct a random hyperparameter search of 60 trials over the space of learning rate \{1e-5, 2e-5, 3e-5\}, epoch limit \{10, 20, 40\}, batch size \{8, 16, 32, 64\}, and random seed.

\paragraph{Datasets}
We run experiments on SuperGLUE WSC and WinoGrande-Medium. We do not cover larger WinoGrande sizes due to computation constraints. Each WSC example includes a sentence, a \textsc{Pron} span, and a \textit{single} marked NP span. Following \citeauthor{liu2019RoBERTa}, for MC-based formalizations, we mine candidate NPs with spaCy\footnote{\url{https://spacy.io/}} and only keep one example from the group of examples that only differ by query NP to avoid redundancy. 

For WinoGrande, each example provides a sentence with two marked NP spans and a fill-in-the-blank to represent the \textsc{Pron}. When using asymmetric formalizations like P-Sent, we duplicate each example, making one option the query and the other the candidate. For P-Span, we use the first appearance of query or candidate NP in the sentence as the NP span and use the blank as \textsc{Pron} span. 

\begin{table}[t]
    \centering
    \resizebox{.45\textwidth}{!}{
    \begin{tabular}{l r r r r r}
        \toprule
        & \multicolumn{3}{c}{WSC} & \multicolumn{2}{c}{WinoGrande} \\
        Formalization   & \multicolumn{1}{l}{Test} & \multicolumn{1}{l}{Std} & \multicolumn{1}{l}{Kurt} & \multicolumn{1}{l}{Std} & \multicolumn{1}{l}{Kurt} \\
        \midrule
        MC-MLM          & 86 & 2  & 3 & 3 & 13\\
        MC-Sent         & 77 & 7  & -1 & 5 & 4 \\
        MC-Sent-NoSoftmax    & 77 & 8  & -1 & 4 & 3 \\
        MC-Sent-NoPairLoss   & 86 & 6 & 5 & 1 & 1 \\
        P-Sent          & 67 & 5 & 24 & 2 & 28 \\
        P-Span          & 80 & 4 & 0 & 7 & -2 \\
        \bottomrule
    \end{tabular}}
    \caption{Validation accuracy standard deviation (\textit{Std}) and excess kurtosis (\textit{Kurt}) for WSC and WinoGrande and test accuracy for WSC using RoBERTa. Test results are from an ensemble of the top five models.}
    \label{tab:stats}
\end{table}

\paragraph{RoBERTa Results}
Figures \ref{subfig:wsc} and \ref{subfig:winogrande} and Table \ref{tab:stats} show the distribution over validation accuracies from 60 runs with each formalization using RoBERTa. We do not report WinoGrande test results since submissions require test set predictions from all five training sets and we only train using WinoGrande-Medium. We also include the majority-class baseline and human performance. From the WSC test results, we find MC-MLM outperforms P-Span. The 6\% gain between P-Sent and MC-Sent-NoPairLoss indicates MC evaluation alone may improve accuracy. However, we also find most formalizations are sensitive to hyperparameter choices and random seeds. Given the small size of the SuperGLUE WSC test set at 146 examples, we find it more informative to focus on the distribution of validation results.

In both datasets, we see three main changes. First, including paired training with MC-Sent-NoSoftmax increases performance variance by adding more weight to the tail of higher scores. Second, we see the weight of higher performances increase even more with softmax scaling in MC-Sent. In WSC, the higher scores become the body of the distribution with smaller variance. In WinoGrande, the distribution of MC-Sent has an increased excess kurtosis indicating the tail of higher scores occur more frequently. Finally, the model achieves higher scores with significantly lower variance using MLM in MC-MLM. This may be a result of fine-tuning the pretrained MLM head rather than a new initialization.

We see two main differences between WSC and WinoGrande results. First, P-Sent performs significantly worse than P-Span on WinoGrande. We suspect this is due to WinoGrande's adversarial filtering removing examples that are easy to classify from sentence representations. Second, MC-Sent-NoPairLoss does not benefit WinoGrande and may indicate the benefit from MC evaluation may not extend to other Winograd like corpora.

\paragraph{BERT Results}
Figure \ref{subfig:bert_results} shows ablation results using BERT and WSC. We find that RoBERTa outperforms BERT with both models using the same MC-MLM formalization, which is in line with leaderboard performances. We also find similar trends across task formalizations in Figure \ref{subfig:wsc}, further highlighting the impact of formalization decisions on performance gains. Most formalizations are still sensitive to hyperparameter choices and random seed, MC evaluation alone provides a benefit over P-Span at the 75\textsuperscript{th} percentile of roughly 6\%, and incorporating MLM provides additional benefits in performance.

However, we also find that using BERT's pretrained MLM head does not provide the lower variance displayed with RoBERTa. Comparing the performances of intermediate formalizations, we see that BERT generally performs worse than RoBERTa. This is consistent with the findings from \citet{tenney2019probing} that show BERT embeddings encode information less suited for coreference resolution during pretraining. Consequently, BERT's pretrained MLM head would be less optimized for a coreference resolution task like WSC than RoBERTa's and may not provide the same stability benefits.

\section{Conclusion}
\label{sec:Conclusion}
By only varying task formalization, we observe a wide range of results among reasonable task formalizations on WSC and WinoGrande evaluations. Having access to candidate NPs during inference alone improves the performance on SuperGLUE WSC. However, models with MC evaluation are highly sensitive to hyperparameters and fail to perform better on WinoGrande. We find training with paired inputs, using a softmax over candidates, and reusing a pretrained MLM head all help to learn commonsense reasoning and reduce this sensitivity. While we find evidence that these formalization choices can largely influence WSC performance, we do not see obvious evidence of similar occurrences on other task comparisons with RoBERTa.

For MC formalizations, we follow \citeauthor{liu2019RoBERTa} for WSC and use spaCy to mine candidate NPs. This extrinsic preprocessing step yields dramatic gains without significantly changing the reasoning ability of the model. We view such gains as orthogonal to the intent of the task and urge benchmark creators to minimize the opportunity for these insubstantial improvements by imposing as much structure as is possible in the released data, for example, by providing candidate NPs explicitly. 

We also encourage future reports of system performances to use the same task formalization whenever possible. At a minimum, greater emphasis should be given to task formalization decisions when they deviate from the prevailing standard. We believe this will help disentangle gains due to models' reasoning abilities, especially in situations where these decisions significantly impact performance, such as in WSC.

Finally, we find that differences between reasonable formalizations can have big impacts on performance with our case study using WSC. For example, using a pretrained MLM task head as the basis for a downstream task classifier yields strong results with very little hyperparameter sensitivity. This echoes the strong results seen with T5 and offers further motivation to explore these kinds of design decisions in other tasks.

\section*{Acknowledgements}
We thank Vid Kocijan, Ethan Perez, Jason Phang, Aishwarya Kamath, Iacer Calixto, Clara Vania, Rasika Bhalerao, Alex Warstadt, Richard Pang, He He, and Ernie Davis for their helpful feedback. This project has benefited from financial support to SB by Eric and Wendy Schmidt (made by recommendation of the Schmidt Futures program), by Samsung Research (under the project \textit{Improving Deep Learning using Latent Structure}), by Intuit, Inc., and in-kind support by the NYU High-Performance Computing Center and by NVIDIA Corporation (with the donation of a Titan V GPU). This material is based upon work supported by the National Science Foundation under Grant No.  1922658. Any opinions, findings, and conclusions or recommendations expressed in this material are those of the author(s) and do not necessarily reflect the views of the National Science Foundation.

\bibliography{main}
\bibliographystyle{acl_natbib}
%\nocite{*}
% The \cite command functions as follows:
%   \citet{key} ==>>                Jones et al. (1990)
%   \citet*{key} ==>>               Jones, Baker, and Smith (1990)
%   \citep{key} ==>>                (Jones et al., 1990)
%   \citep*{key} ==>>               (Jones, Baker, and Smith, 1990)
%   \citep[chap. 2]{key} ==>>       (Jones et al., 1990, chap. 2)
%   \citep[e.g.][]{key} ==>>        (e.g. Jones et al., 1990)
%   \citep[e.g.][p. 32]{key} ==>>   (e.g. Jones et al., p. 32)
%   \citeauthor{key} ==>>           Jones et al.
%   \citeauthor*{key} ==>>          Jones, Baker, and Smith
%   \citeyear{key} ==>>             1990

% \input{writing/appendices}

\end{document}